\pdfoutput=1

\documentclass[11pt]{article}

\usepackage{EMNLP2023}

\usepackage{times}
\usepackage{latexsym}
\usepackage{amsmath}
\usepackage{multicol}
\usepackage{multirow}
\usepackage{graphicx}
\usepackage{longtable}
\usepackage{tabularx}
\usepackage{balance}
\usepackage{booktabs}
\usepackage{tcolorbox}
\usepackage{xcolor}
\usepackage{float}
\usepackage{enumitem}
\usepackage{verbatim}
\usepackage{listings}

\lstdefinelanguage{Kusto}{
  keywords={let, in, where, project, summarize, by},
  morekeywords=[2]{count, distinct, toscalar, union},
  morekeywords=[3]{|, as, on, with},
  sensitive=true,
  morecomment=[l]{//},
  morecomment=[s]{/*}{*/},
  morestring=[b]"
}

\definecolor{lightgray}{RGB}{220, 220, 220}
\definecolor{darkblue}{RGB}{0, 0, 128}

\newtcolorbox{promptbox}{
    colback=lightgray,
    colframe=lightgray,
    arc=4pt,
    fontupper=\sffamily\small,
    colupper=darkblue,
    boxrule=0.5pt,
    sharp corners
}

\usepackage[T1]{fontenc}

\usepackage[utf8]{inputenc}

\usepackage{microtype}

\usepackage{inconsolata}
\usepackage{xspace}

\newcommand{\eg}{\emph{e.g.}\xspace}
\newcommand{\ie}{\emph{i.e.}\xspace}
\newcommand{\etaldot}{\emph{et al.}\xspace}

\title{Empower Large Language Model to Perform Better\\ on Industrial Domain-Specific Question Answering}

\author{Fangkai Yang$^1$ \quad  Pu Zhao$^1$  \quad Zezhong Wang$^2$\thanks{\quad Work done during the internship at Microsoft.} \quad  Lu Wang$^1$\\ \quad {\bf Jue Zhang}$^1$ \quad {\bf Mohit Garg}$^1$\quad {\bf Qingwei Lin}$^1$ \quad {\bf Saravan Rajmohan}$^1$ \quad {\bf Dongmei Zhang}$^1$    \\
      $^1$Microsoft\\ $^2$The Chinese University of Hong Kong\\
      }

\begin{document}
\maketitle
\begin{abstract}
Large Language Model (LLM) has gained popularity and achieved remarkable results in open-domain tasks, but its performance in real industrial domain-specific scenarios is average due to its lack of specific domain knowledge. This issue has attracted widespread attention, but there are few relevant benchmarks available. 
In this paper, we provide a benchmark Question Answering (QA) dataset named MSQA, centered around Microsoft products and IT technical problems encountered by customers. This dataset contains industry cloud-specific QA knowledge, an area not extensively covered in general LLMs, making it well-suited for evaluating methods aiming to enhance LLMs' domain-specific capabilities. In addition, we propose a new model interaction paradigm that can empower LLM to achieve better performance on domain-specific tasks where it is not proficient.
Extensive experiments demonstrate that the approach following our method outperforms the commonly used LLM with retrieval methods. We make our source code and sample data available at: \url{https://aka.ms/Microsoft_QA}.

\end{abstract}

\section{Introduction}

Recent advancements in large language models (LLMs), including OpenAI's GPT-3.5~\cite{ouyang2022training}, GPT-4~\cite{OpenAI2023GPT4TR}, Google's PaLM~\cite{chowdhery2022palm}, and other benchmark models~\cite{touvron2023llama, alpaca, Vicuna}, have demonstrated impressive performance across various natural language processing (NLP) tasks. These models are pretrained on extensive data, which imbues them with remarkable language understanding and generation capabilities~\cite{bubeck2023sparks}. However, when it comes to domain-specific problems, LLMs exhibit limited performance due to their insufficient pretraining on domain knowledge, where the overwhelming presence of domain-general data causes them to prioritize common knowledge, leading to a potential oversight of crucial domain-specific information~\cite{lee2023benefits,castro2023large,lecler2023revolutionizing}. Fine-tuning and maintaining LLMs to incorporate domain-specific knowledge can be expensive for most companies and researchers. Moreover, the availability of domain-specific data is often restricted and confidential, introducing risks of potential data leakage during fine-tuning of LLMs~\cite{Samsung}.

\begin{figure*}[!tb]
\centering
     \includegraphics[width=0.98\textwidth]{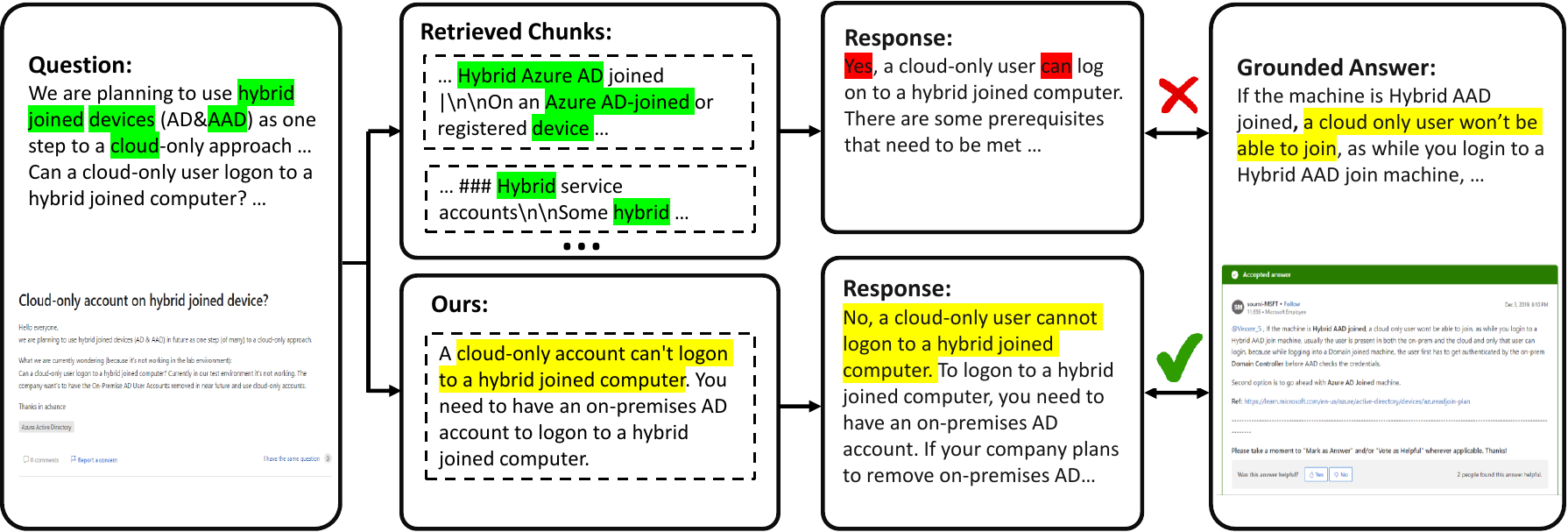}
     \caption{An example\protect\footnotemark from MSQA dataset shows retrieval-based methods' limitations in complex question handling. The retrieved chunks contain matching keywords (highlighted in green) but failed to retrieve essential information needed to answer the question correctly. Our model generates more accurate answers by understanding the question and leveraging domain-specific knowledge (highlighted in yellow). Case details are in Table~\ref{tab:frontcasedetail} in Appendix~\ref{appendix:casestudy}.}
     \label{fig:data_retrieval_example}
     \vspace{-4mm}
\end{figure*}

The existing works primarily focus on enhancing the performance of LLMs in specific domains by employing retrieval-based methods \cite{Liu_LlamaIndex_2022,shi2023replug,peng2023check} or external modules \cite{wu2023visual,AutoGPT} to extract domain-specific knowledge. However, these approaches suffer from certain limitations. Firstly, retrieval-based methods face challenges in handling complex queries as they may not retrieve all the necessary domain-specific context, leading to incomplete information.  
Additionally, retrieved chunks can suffer from the issue of ``quote out of context'' where the retrieved information may not fully capture the intended meaning~\cite{engel1982good}.
As the example shown in Figure~\ref{fig:data_retrieval_example}, retrieved chunks contain keywords or spans of the question, but they do not understand the question, resulting in generating inaccurate answers.
Secondly, due to the length limitation of prompts, it becomes challenging to incorporate all retrieved data for in-context learning. This poses a constraint on these methods in leveraging domain-specific knowledge.

Humans integrate domain-specific knowledge with domain-general knowledge through interactions \cite{siegler1989domain,penner1996interaction,li2014domain}. For example, Penner \etaldot~\cite{penner1996interaction} conducted an experiment where children inferred factors affecting the sinking rates of objects. Initially, children believed weight alone determined sinking, but the experiment helped them understand the effects of object shape and material on sinking rates.
This domain-specific knowledge was extracted and learned through interactive experiences with various objects, rather than being conveyed through formal, abstract rules. 
Inspired by this, we introduce a novel model interaction paradigm that bridges domain-general and domain-specific knowledge. Our approach involves fine-tuning a smaller LLM, \ie, LLaMA~\cite{touvron2023llama}, using domain documentation to align it with domain-specific knowledge. At runtime, our fine-tuned model provides domain-specific knowledge to LLMs. This paradigm replaces traditional retrieval modules with the generation of domain-specific knowledge, enabling easy maintenance and privacy protection within the specific domain.

\footnotetext{The QA example details can be found in \url{https://learn.microsoft.com/en-us/answers/questions/2096/}}

In this paper, we focus on the cloud domain and specifically address question-answering (QA) tasks using our proposed model interaction paradigm. While LLMs have demonstrated their effectiveness in QA tasks, there is limited exploration and evaluation of LLMs in domain-specific QA tasks involving long-form answers. Our contributions are summarized as follows: 

\begin{itemize}[noitemsep,topsep=0pt]
\item We release a cloud-domain QA dataset that contains 32k QA pairs from the Microsoft Q\&A forum\footnote{The data is collected and post-processed from the Microsoft Q\&A forum (\url{https://learn.microsoft.com/en-us/answers/questions/}), which is publicly available.}. To the best of our knowledge, this is the first cloud-domain QA dataset. We believe that this benchmarking dataset will assist the research community in evaluating their models in domain-specific scenarios.
\item We propose a new model interaction paradigm that empowers the LLM with generated domain-specific knowledge. Evaluation results highlight the significant performance of our model interaction paradigm in generating answers enriched with domain-specific knowledge, compared with retrieval-based methods.

\item We propose novel evaluation metrics for assessing long-form answers in QA tasks, which are aligned with human evaluations and have the potential for automation evaluation.

\end{itemize}

\section{Related Work}

\subsection{Question Answering Datasets}
Question answering (QA)~\cite{hirschman2001natural} aims to provide answers based on knowledge or given context.
Recent advancements in LLMs have shown promising results in various QA datasets~\cite{wang2022modern}. However, existing evaluations mainly focus on answer types like multiple-choice or span extraction, which are comparatively easier to assess LLM performance. Evaluating long-form question answering (LFQA)~\cite{fan2019eli5,krishna2021hurdles,nakano2021webgpt,su2022read} poses challenges due to limited datasets and appropriate evaluation metrics.
In particular, LLMs are often not evaluated in specific domains, and available domain-specific QA datasets, such as medical~\cite{pal2022medmcqa,jin2019pubmedqa}, financial~\cite{chen2021finqa}, and legal domains~\cite{zheng2021does}, typically include questions, answers, and relevant paragraphs. However, in practical QA scenarios, this additional contextual information may not always be available. Our paper addresses this by releasing an LFQA dataset specific to the cloud domain, along with new evaluation metrics. Our approach eliminates the need for an additional paragraph to extract domain-specific knowledge, making it suitable for industrial applications while ensuring data privacy.

\subsection{Augmented Large Language Models}

Recent efforts have been made to enhance the context generation ability of LLMs in specific domains by incorporating external knowledge~\cite{mialon2023augmented}.
One group of approaches leverages external modules, such as
Visual ChatGPT~\cite{wu2023visual}, 
HuggingGPT~\cite{shen2023hugginggpt} 
and Auto-GPT~\cite{AutoGPT}. They highly rely on the LLM's prompting management and the availability of external tools or applications. However, such external modules are not always available when it comes to domain-specific scenarios.
Another group of approaches is retrieval-augmented~\cite{Liu_LlamaIndex_2022,guu2020retrieval,izacard2022few,shi2023replug}, which leverages retrieval-based methods like BM25~\cite{robertson2009probabilistic} and dense passage retrieval (DPR)~\cite{karpukhin2020dense}. 
This approach retrieves relevant data or text chunks, which are then used as additional context to incorporate domain-specific knowledge with LLMs, thus improving their performance.
However, they may not be able to handle complex questions that require information from multiple sources or modalities. Our method is able to comprehend complex questions and provide comprehensive domain-specific knowledge without the ``quote out of context'' issue.

\section{MSQA Dataset Creation}
Current public Q\&A forums, such as Quora, Reddit, Stack Overflow, contain responses to a variety of open-ended questions. However, there are limited Q\&A forums dedicated to specific domains that have a large number of active users. In light of this, we chose to focus on the publicly available Microsoft Q\&A forum\footnote{\url{https://learn.microsoft.com/en-us/answers/}} for our dataset creation, primarily due to its extensive collection of available questions and corresponding answers. These domain-specific QAs cover a wide range of Microsoft technologies and products, such as Azure and Microsoft 365. Additionally, Microsoft offers publicly available and well-documented documentation, which serves as a valuable external resource for extracting domain-specific knowledge. We make our MSQA dataset openly accessible to the NLP community. We hope this resource could facilitate the exploration of LLM's capabilities in handling industrial domain-specific questions.

\subsection{Data Collection and Post-Processing}
We select questions and answers spanning from the Microsoft Q\&A forum from October 2019 to May 2023. These QA pairs went through a filtering process based on user ratings. 
Firstly, we retain QA pairs where the answers were marked as `Accepted'.
Secondly, we exclude QA pairs involving multi-turn discussions, as they are outside the scope of this paper. Additionally, we focus on text-based QA pairs and discard samples containing image attachments, leaving multi-modality QA tasks for future work. Furthermore, we gather metadata of each QA pair, including the number of up-votes received by both the question and answer, question tags, and other relevant information.

The QA pairs obtained through the aforementioned collection process may contain noisy information, particularly in human-written answers. This noise stems from the inclusion of irrelevant details like user names, IDs, decoration symbols, and platform-generated context. They introduce unwanted noise during the fine-tuning process. To mitigate this, we conduct additional data post-processing, following a set of principles detailed in Appendix~\ref{appendix:datafilteringandpostprocessing}. 

\subsection{Statistics}

Following data post-processing, our dataset consists of 32k QA pairs. Table~\ref{tab:dataset_statistics} summarizes the statistics. Each question within the dataset is accompanied by a diverse range of relevant topic tags, comprising a total of 332 distinct tags, such as \textit{Azure Virtual Machine, PowerPoint, Windows Server}. These tags serve to categorize and provide contextual information for the questions. To gain a preliminary understanding of the different types of questions, we employ a categorization approach based on the first interrogative words. The majority of questions fall into the ``Is'' category, which seeks judgments (\textit{Is it possible to ...}), while others require explanations from answers, such as ``How'' or ``Why''. Interestingly, even ``Is'' questions often elicit explanatory answers. Table~\ref{tab:dataset_example} in Appendix~\ref{appendix:statisticsandqsamples} shows randomly sampled examples of MSQA questions based on their types.

\begin{table}[]
\begin{tabular}{ll|lc}
\toprule
\multicolumn{2}{c|}{\textbf{Question Tag (\%)}}        & \multicolumn{2}{c}{\textbf{$1^{st}$ Question word (\%)}} \\ \hline
Azure                   & 28.55               & Is                       & 19.18                    \\
Windows                 & 16.73               & How                      & 11.91                    \\
M365                    & 15.14               & Why                      & 10.75                    \\
Other                   & 39.58               & Do                       & 7.14                     \\ \cline{1-2}
\multicolumn{2}{c|}{\textbf{Avg \# of token}} & Can                      & 6.57                     \\ \cline{1-2}
Question                & 347.15              & What                     & 5.94                     \\
Answer                  & 382.18              & Other                    & 38.33                    \\ \bottomrule
\end{tabular}
\vspace{-2mm}
\caption{Statistics of MSQA}
\label{tab:dataset_statistics}
\vspace{-4mm}
\end{table}

\section{Methodology}
\begin{figure}[tb]
\centering
     \includegraphics[width=0.48\textwidth]{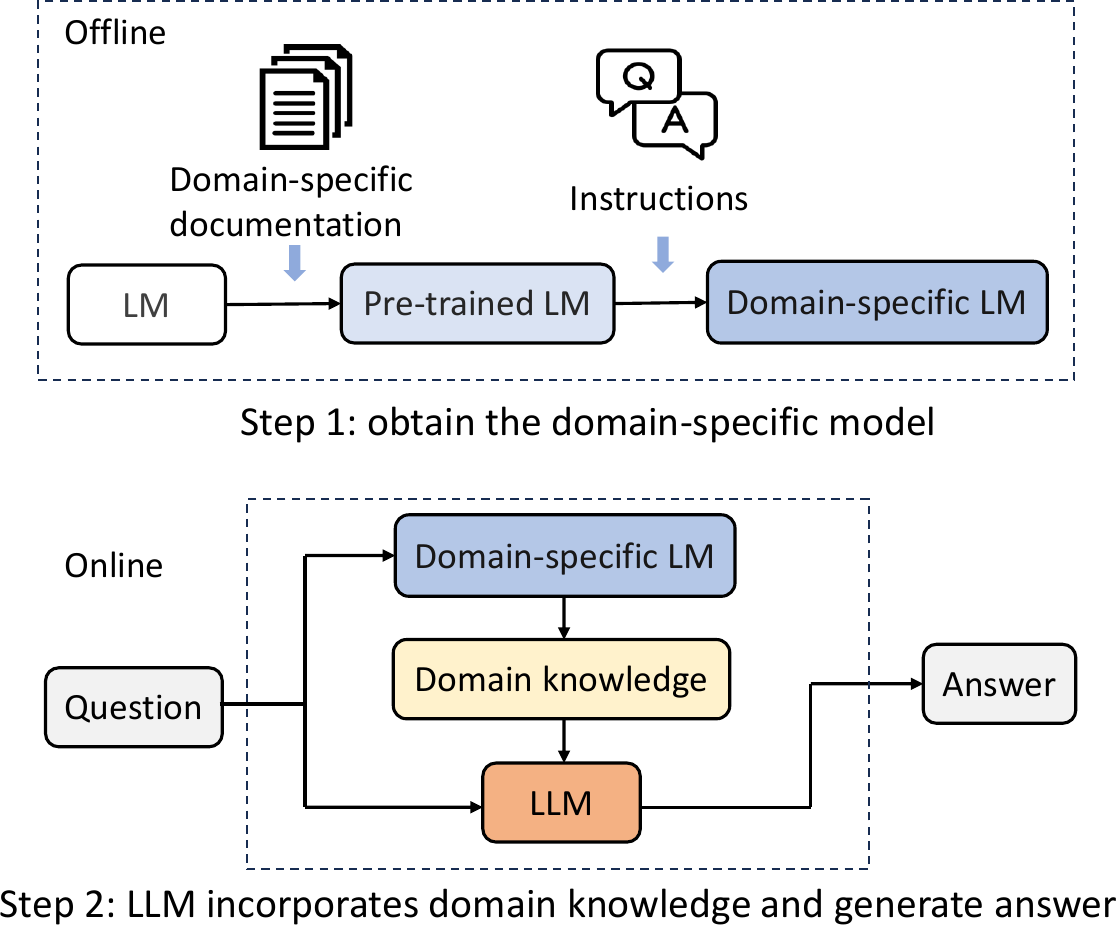}
     \caption{The model interaction framework.}
     \label{fig:framework}
     \vspace{-4mm}
\end{figure}

The model interaction paradigm (shown in Figure~\ref{fig:framework}) involves two key steps: (a) obtaining a domain-specific model that incorporates aligned knowledge, (b) providing the generated domain-specific knowledge to LLMs, enabling them to generate answers enriched with domain knowledge. 

In the first step, we pre-train small language models\footnote{We use LLaMA-7B~\cite{touvron2023llama} in this paper.} using the publicly available Azure documentation\footnote{\url{https://github.com/MicrosoftDocs/azure-docs}}. This documentation serves as a comprehensive knowledge base for Microsoft Azure, containing cloud-domain knowledge of Microsoft's cloud products. Note that Microsoft maintains extensive documentation covering various product offerings. However, we specifically focus on Azure documentation as it aligns with the prevalent tags related to Azure found in the Q\&A forum, which captures the most frequently asked questions (shown in Table~\ref{tab:dataset_statistics}). By narrowing down our focus to Azure, we aim to evaluate the efficacy of our model interaction paradigm within a well-defined domain.

After completing the pre-training phase, we imbue the small language models with domain-specific knowledge from the Azure knowledge base. We then adapt the model to the LFQA task through instruction tuning~\cite{ouyang2022training}, allowing it to specialize and become more accurate in the QA task.
To facilitate instruction tuning, we construct instructions from the training set of the MSQA dataset. Each instruction consists of a three-element tuple, including an instruction prompt, an input query or statement, and a corresponding response. The instruction template is shown in Table~\ref{tab:instructiontemplate} (Appendix~\ref{appendix:instructiontuning} shows an example instruction). The details of the training setup and parameters can be found in Appendix~\ref{appendix:trainingparameter}.

\begin{table}[H]
\small
\begin{tcolorbox}
Below is an instruction that describes a task. Write a response that appropriately completes the request.\\
\textbf{Instruction}: Please answer the following questions concerning \textcolor{blue}{\textbraceleft Tags\textbraceright}.\\
\textbf{Input}: \textcolor{blue}{\textbraceleft Question\textbraceright}\\
\textbf{Response}: \textcolor{blue}{\textbraceleft Answer\textbraceright}
\end{tcolorbox}
\vspace{-3mm}
\caption{The instruction template.}
\label{tab:instructiontemplate}
\vspace{-5mm}
\end{table}

By engaging in instruction tuning, the pre-trained small LM learns and assimilates domain-specific knowledge, enabling it to generate relevant responses when encountering domain-specific queries or statements.

In the second step, the fine-tuned domain-specific language model serves as an expert in Azure. During runtime, the domain-specific LM leverages its knowledge to provide domain-relevant information in response to a given question. Then the LLM takes both the question and the domain knowledge to generate the final response. 
By enriching the LLMs with domain-specific knowledge, their comprehension of the question context is enhanced, resulting in more accurate and contextually appropriate responses.
Note that our approach does not propose replacing the LLM with a domain-specific LM. Instead, we propose a model interaction paradigm, leveraging the domain-specific LM as an expert to provide knowledge. Through our application practice, we have observed that domain-specific knowledge may not excel in language expression and general question answering, as questions may contain both Azure-related and general queries. Additionally, our domain-specific model can function as a compatible plugin within the existing retrieval-based system, offering supplementary information beyond just chunks.

\section{Experiment}

\subsection{Baselines}

We leveraged two LLMs, namely GPT-3.5 (\texttt{gpt-35-turbo}) and GPT-4 (\texttt{gpt-4}), as the backbone to output the answer by taking the extra information from either the data-retrieval methods or our approach.
We utilize two data retrieval methods, \ie, BM25~\cite{robertson2009probabilistic} and dense passage retrieval (DPR)~\cite{karpukhin2020dense}. These methods were employed to retrieve the top-3 relevant information chunks from Azure documentation, which were then used as supplementary information for the backbone LLMs during answer generation. We make the below baselines:

\noindent\textbf{Raw LLM (LLM)}. Questions were directly posed to the backbone LLMs without providing any additional information.

\noindent\textbf{LLM+BM25/+DPR}. The LLM incorporated both the question and retrieved chunks using BM25 and DPR, respectively.

\noindent\textbf{LLM+EXP}. The LLM utilized the domain knowledge from our domain-specific LM as extra information to generate answers.

Appendix~\ref{appendix:baselineprompt} shows the baseline prompt details.

\subsection{Evaluation Metrics}\label{subsec:metrics}

Evaluating long-form generated answers lacks an automatic metric, and thus, we employ standard metrics, our proposed metrics, and human evaluation to assess the quality of the generated answers.

\noindent\textbf{Lexical-Overlap-Based Metrics}. We employ BLEU~\cite{papineni-etal-2002-bleu}, ROUGE-1, ROUGE-2, and ROUGE-L~\cite{lin-2004-rouge}, and METEOR~\cite{banerjee2005meteor}, as the lexical-overlap-based metrics to measure the N-gram alignment between the generated answers and the grounded answers. 

\noindent\textbf{Semantic-Overlap-Based Metrics}. To evaluate the semantic overlap between the generated answers and the ground truth, we utilize BERT-Score~\cite{zhang2020bertscore} 
and report F1 score. Additionally, we calculate the similarity between the embedding of the grounded answer and the embedding of the generated answer, referred to as the SIM metric.

Besides the above metrics, we propose three novel metrics for the LFQA scenario: 
\noindent\textbf{Keyword/Span-Hit-Rate (KHR)}. We extract keywords or spans from the grounded answer, removing those presented in the question. This yields a distinct keyword/span set, and we measure the rate of hits in the generated response (Table~\ref{tab:khr} in Appendix~\ref{appendix:baselineprompt} shows the prompt). 

\noindent\textbf{Can-Answer-Rate (CAR)}. To prevent answer hallucinations, we require the backbone LLMs to answer only when confident. CAR represents the percentage of answerable questions and evaluates the informativeness of extra information provided by data-retrieval methods or our approach.

\noindent\textbf{LLM-based Metrics}. LLMs have demonstrated impressive  performance as evaluators and annotators in recent studies~\cite{wang2022self, chiang2023vicuna, peng2023instruction}. In our work, we employ an LLM as an evaluator to compare and rank two responses based on their similarity to the grounded answer (see full prompt in Appendix~\ref{appendix:evaluatorprompt}).
 However, concerns have been raised regarding the reliability of LLMs as evaluators due to their sensitivity to response positions~\cite{wang2023large}. To address this issue, we incorporate the chain-of-thought concept~\cite{wei2022chain} in our prompt design, which involves providing detailed explanations before scoring the responses.
  Moreover, we propose a rule where we trust the LLM evaluator only when the score gap exceeds 1 (excluding 1), allowing for a single round of scoring. Otherwise, we conduct two scoring rounds, switching response positions, and rank them based on the average score of the two rounds. Note that GPT-4 exhibits significantly fewer conflict cases compared to GPT-3.5, leading us to select GPT-4 as the evaluator. Further details of the score gap study are available in Appendix~\ref{appendix:evaluatorsensitity}.

\noindent\textbf{Human Evaluation}. There still requires human evaluation as there is a lack of good metrics of long-form answers~\cite{fan2019eli5,krishna2021hurdles}. We evaluate a small subset of test samples (30 randomly sampled QA pairs). Five evaluators with domain knowledge are given QA pairs and three responses from different methods. They are asked to rank these three responses based on their similarity with the grounded answer. 
The evaluation setup and user interface are in Appendix~\ref{appendix:humanevaluation}.

\section{Results}
As suggested in \cite{krishna2021hurdles, ji2023survey} and our experiments, the lexical-overlap-based metrics are not an informative way to evaluate the quality of LLM-generated answers due to their poor correlation with grounded human-written answers. As shown in Table~\ref{tab:lexicalmetric35} and~\ref{tab:lexicalmetric4} in Appendix~\ref{appendix:lexicalresults}, the lexical-overlap-based scores demonstrate fewer variations across different methods, and the scores are low in general, suggesting that these metrics are not suited.

\begin{table}[!tb]
\centering\resizebox{0.49\textwidth}{!}{
\begin{tabular}{l|cccc}
\toprule

\textbf{Metrics (\%)}   & \textbf{LLM} & \textbf{LLM+BM25} & \textbf{LLM+DPR} & \textbf{LLM+EXP} \\ \hline
\textbf{BERT-Score}                & 52.47 & 53.83      & 54.94  & \textbf{56.21} \\
\textbf{SIM }                 & 61.84   & 62.46      & 64.87      & \textbf{67.08}    \\
\textbf{KHR}                 &   22.53   & 23.25   &  24.30     & \textbf{24.61}    \\
\textbf{CAR}                 & 98.37  & 92.07      & 95.34     & \textbf{99.77}     \\ \bottomrule
\end{tabular}
}
\vspace{-2mm}
\caption{The results of semantic-overlap-based metrics over different methods with GPT-3.5 as backbone LLM.}
\label{tab:result35}
\vspace{-3mm}
\end{table}

\begin{table}[!tb]
\centering\resizebox{0.49\textwidth}{!}{
\begin{tabular}{l|cccc}
\toprule

\textbf{Metrics (\%)}   & \textbf{LLM} & \textbf{LLM+BM25} & \textbf{LLM+DPR} & \textbf{LLM+EXP} \\ \hline
\textbf{BERT-Score}                & 51.79  & 52.33       &  54.83      & \textbf{56.91}      \\
\textbf{SIM}                 & 67.94  & 68.30       &68.78      &  \textbf{71.19}      \\
\textbf{KHR}                 &  30.40  & 32.15    &   32.50  & \textbf{33.13 } \\
\textbf{CAR}                 & 76.22  & 73.89       & 87.41      & \textbf{99.30}      \\ \bottomrule
\end{tabular}
}
\vspace{-2mm}
\caption{The results of semantic-overlap-based metrics over different methods with GPT-4 as backbone LLM.}
\label{tab:result4}
\vspace{-3mm}
\end{table}

Table~\ref{tab:result35} and~\ref{tab:result4} show the results of semantic-overlap-based metrics, \ie BERT-Score and SIM, with GPT-3.5 and GPT-4 serving as the backbone LLMs for answer generation, respectively. The worst performance is observed for Raw LLM, highlighting the usefulness of extra information provided through data-retrieval methods or our method. LLM+DPR has better performance than LLM+BM25, and our LLM+EXP achieves the best performance. Note that the difference between Raw LLM and other baselines is relatively small, possibly due to the pre-training of LLMs, which already contains some knowledge related to Microsoft Azure. Our KHR metric has a similar pattern as the lexical-overlap-based metric. However, we observe that CAR is initially high for Raw LLM with GPT-3.5 (Table~\ref{tab:result35}), but decreases when extra information from data-retrieval methods is provided. This suggests that GPT-3.5 may exhibit blind confidence, leading to potential answer hallucinations. By incorporating extra information, it gains access to relevant information and is not solely reliant on its own knowledge. In contrast, GPT-4 demonstrates superior performance and is not blindly confident in its answers, even without extra information (76.22\% CAR in Table~\ref{tab:result4}). Note that responses that cannot answer the question, \eg, \textit{``Sorry, I cannot give a confident answer.''}, are excluded when calculating other metrics.

LLM+DPR performs better than LLM+BM25, as indicated by the previous analysis. Hence, we select LLM+DPR as the representative data-retrieval method for both LLM-based metric evaluation and human evaluation to optimize resources and reduce human efforts. In the LLM-based metric evaluation, we compare methods pairwise three times and exclude samples with circular preferences or rank conflicts (17.97\% conflict rate over the test set). Table~\ref{tab:llmevaluator} demonstrates that LLM+EXP outperforms baselines, achieving the highest favor rate and the average rank. The favor rate means the percentage of a certain method selected as the best over the test set.
Table~\ref{tab:humaneval} shows the human evaluation result with at least two agreements among evaluators. Similar to the LLM-based metric, LLM+EXP shows the best performance in the favor rate and the average ranking. Moreover, LLM+EXP has the least ``Don't Know'' rate, representing the confidence of the human evaluators. The agreement analysis in Appendix~\ref{appendix:evaluationagreements} shows that human evaluation is reliable and consistent among evaluators. The results align with the LLM-based metric, highlighting the significant performance of our method and the potential of using the LLM-based metric as an automation evaluation.
We present case studies in Appendix~\ref{appendix:casestudy} to give a comprehensive comparison of different methods. The retrieved-based methods tend to provide scattered and often ``quote out of context'' chunks. In contrast, the domain knowledge from our method offers more concise and relevant information, with a significantly shorter length compared with the retrieved chunks.

\begin{table}[!tb]
\centering\resizebox{0.43\textwidth}{!}{
\begin{tabular}{l|ccc}
\toprule
                         & \textbf{LLM} & \textbf{LLM+DPR} & \textbf{LLM+EXP} \\ \hline
\textbf{Most Favor (\%)} &  51.98 & 52.45   & \textbf{68.76}  \\
\textbf{Avg Rank}        &  1.33  & 1.29   & \textbf{1.05} \\ \bottomrule
\end{tabular}
}
\vspace{-2mm}
\caption{The results of LLM-based metric. Ranks: 1 (highest), 2 (second), and 3 (lowest). Ranks can be tied.}
\label{tab:llmevaluator}
\vspace{-2mm}
\end{table}

\begin{table}[!tb]
\centering\resizebox{0.43\textwidth}{!}{
\begin{tabular}{l|ccc}
\toprule
                         & \textbf{LLM} & \textbf{LLM+DPR} & \textbf{LLM+EXP} \\ \hline
\textbf{Most Favor (\%)} &  13.33   &    20.00    &   \textbf{76.67 }        \\
\textbf{Avg Rank}        &  2.19  &  2.07      &  \textbf{1.34}          \\ 
\textbf{Don't Know}        &  0.13  &  0.10       &         \textbf{0.03} \\ \bottomrule
\end{tabular}
}
\vspace{-2mm}
\caption{The results of human evaluation.}
\label{tab:humaneval}
\vspace{-4mm}
\end{table}

\section{Conclusion}
In this paper, we deal with the challenge of empowering LLMs with domain-specific knowledge, enabling them to accurately answer questions in industrial scenarios. Due to the limited availability of relevant benchmarks, we introduce the MSQA dataset, tailored for cloud domain QA. Our novel model interaction paradigm effectively equips LLMs with domain-specific knowledge, bridging the gap between general models and industry demands. 
Experiments demonstrate and highlight the effectiveness of our proposed paradigm in standard and newly proposed metrics.

\newpage
\section*{Limitations}
It is essential to discuss the limitations of this paper. One primary limitation is the dataset used for experimentation is confined to Microsoft Azure. It potentially impacts the generalizability of the proposed model interaction paradigm in other domain-specific scenarios. Another limitation is the parameter tuning in instruction tuning. It is unlike pre-training, where we have a large amount of data to perform a few epochs to make the model imbue Azure domain knowledge. In instruction tuning, it is challenging to set the number of epochs properly. There still lacks a well-defined and automated metric to evaluate LFQA in order to select good checkpoints with less effort. From our practice, setting a large max token length and more epochs does not necessarily make a better model. Moreover, this paper focuses on text-based QA, excluding QA scenarios with image attachments. Lastly, the proposed model is trained and evaluated exclusively in English, while the Microsoft Q\&A forum includes QAs in other languages. These limitations constrain the applicability of our model to other languages and multi-modality scenarios.



\section*{Ethics Statement}
Although we use language models trained on data collected from the web, which have been shown to have issues with gender bias and abusive language, we have taken significant precautions to ensure the ethical integrity of our research. Our pre-training and instruction-tuning data have been carefully verified to exclude any toxic text, and we collected the data from the Microsoft Q\&A forum, where engineering experts and administrators take moderation and inspection. We have implemented rigorous filtering mechanisms and conducted thorough validation to remove inappropriate content and any user information. All data used, including human evaluation data, is anonymized and processed in compliance with privacy regulations, with no disclosure of personally identifiable information. While acknowledging the limitations and the need for ongoing research, we are dedicated to advancing responsible and unbiased AI technologies and welcome any inquiries regarding the ethical aspects of our work.

\bibliography{custom}
\bibliographystyle{acl_natbib}

\appendix
\section{Data Post-Processing}\label{appendix:datafilteringandpostprocessing}
Due to the fact that the data is collected from an online Q\&A forum, the context of answers is usually complex and includes a large number of decorative symbols and platform-generated content, which makes the data not easy to use and causes potential noise in fine-tuning.
To address this issue, we conducted a deep sampling of the collected data in order to summarize the existing problems and identify their patterns. We design the following data filtering pipeline:
\begin{itemize}
    \item Remove user-related information, such as usernames and IDs, \eg, \texttt{bob@1234567}, as these personal details are irrelevant to the QA content and contain noise. For example, including such information in the instruction-tuning data would make fine-tuned model output answers starts with hallucinated user name and IDs. Additionally, removing this information helps protect privacy.
    \item Standardize all links appearing in the data according to the Markdown link reference syntax, organizing them into a unified format, \ie, \texttt{[description](link)}. We find these links are also meaningful, and the model could extract information from the context of the links. The fine-tuned model generates relevant and valid links in the response.
    \item Remove platform-generated contents, such as
\begin{quote}
    \texttt{"--please don't forget to upvote and Accept as answer if the reply is helpful--"}    
\end{quote}
\item  Remove irregular decorative symbols added by users, such as \texttt{"****"} for separation.
\item  Address different types of line breaks and handling consecutive multiple line breaks. We adopted a strategy to replace consecutive multiple line breaks with a single line break, while preserving the integrity of code blocks by not merging multiple spaces within them.
\item Detect image links in questions and remove samples with screenshots. This dataset focuses solely on plain text, with multimodal data exploration reserved for future work.
\item Detect the length of questions and specifically label samples with questions exceeding 8192 tokens, as these may require special handling or truncation for current models.
\end{itemize}

\begin{figure}[!htb]
\centering
     \includegraphics[width=0.48\textwidth]{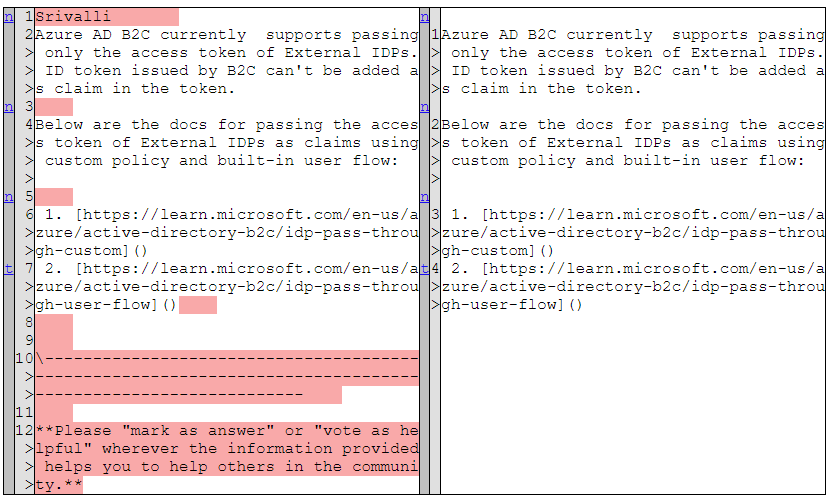}
     \caption{An example of data post-processing. The highlighted red part is removed in post-processing.}
     \label{fig:dataprocess}
\end{figure}

For reference, Figure~\ref{fig:dataprocess} provides an example of data post-processing, showcasing the removal of user names and platform-generated context.

By implementing this data filtering pipeline, we aim to improve the quality and usability of the collected data for research purposes.

\section{Statistics and Question Samples}\label{appendix:statisticsandqsamples}

As shown in Table~\ref{tab:statisticsdetailed}, the average length of questions is 347.15 tokens, whereas the average length of answers is 382.18 tokens. Additionally, based on the analysis conducted, the average upvote counts are 0.05 for questions, 0.26 for answers, and 0.32 for samples. Upvotes serve as an indicator of the content's perceived value by other users within the community, and these counts have been collected and calculated independently.

\begin{table}[h]
\centering
\begin{tabular}{lc}
\hline
\textbf{Statistic} & \textbf{Value} \\
\hline
\#data & 32252 \\
\#tags & 332 \\
Avg. \#questions per tag & 97.36 \\
Avg. \#tags per question & 1.35 \\
Avg. \#tokens per question & 347.15 \\
Avg. \#tokens per answer & 382.18 \\
Avg. \#upvotes per question & 0.05 \\
Avg. \#upvotes per answer &  0.26 \\
Avg. \#upvotes per sample & 0.32 \\
\hline
\end{tabular}
\caption{Additional statistics of MSQA}
\label{tab:statisticsdetailed}
\end{table}

We randomly sampled questions based on their types as shown in Table~\ref{tab:dataset_example}.

\begin{table*}
\begin{tabular}{p{0.9\linewidth}}
\toprule
\textbf{Is}\\
Looking at migration and also backup/restore options for ConfigMgr. Historically, Microsoft do not support either of the below for ConfigMgr Primary servers:
\begin{itemize}
    \item **Migration to VM method**; Physical to Virtual (P2V)
    \item **Backup method**; VM snapshots
\end{itemize}
\colorbox{lightgray}{Is that still the case for both these scenarios?}\\
\hline
\textbf{How}\\
\colorbox{lightgray}{How to have administrator reset password for ADB2C user?} I'm trying to reset passwords for users inside of the Azure ADB2C blade but when trying the temporary password I always get "The password has expired" error message.\\
\hline
\textbf{Why}\\
Password Writeback General Questions. So, I'm trying to understand some more intricate workings of PasswordResetService. Unlike the pass-through authentication feature, there is no Windows service that runs for password writeback. It is my understanding that password writeback uses a service bus relay that's specific to the tenant. \colorbox{lightgray}{Why do I not see that in my tenant,} and how is this working under-the-hood? Is there no need for multiple "instances" like there is for pass-through Authentication? Is it a WCF service, and if so, what is that doing, and how is high availability accounted for?\\
\hline
\textbf{Do}\\
I changed my app service plan level and that led to a change of the inbound IP address. Now I have several apps running there where the domain is handled externally. I had no problems changing the A record for these.

However, I also have an "app service domain" managed by Azure pointing there. \colorbox{lightgray}{Do I have to do anything for this domain, or is the change propagated automatically?} If I have to do something, where do I find the documentation, because I can't find any.\\
\hline
\textbf{Can}\\
For my Windows 11 laptop, \colorbox{lightgray}{can I use the same Windows product key on my VMs, without} \colorbox{lightgray}{having to pay a license for each VM?}\\
\hline
\textbf{What}\\
We have a Hybrid Exchange environment, and many user mailboxes are still on-prem. However, many users already have an E3 license. The issue occurs when a user is logged into OWA and is connecting to our on-premise exchange server. When they receive an O365 link (e.g., [https://forms.office.com/](https://forms.office.com/)), the user gets the error, 'You do not have permission to view or respond to this form', even though the user has an E3 license. When they open up an incognito window and sign into O365 with the same credentials, everything works flawlessly. If someone could explain the theory behind how this works, that would be great. \colorbox{lightgray}{What is the difference between these two credentials even though the credentials are exactly} \colorbox{lightgray}{the same?} Thank you.\\
\bottomrule
\end{tabular}
\caption{Examples of questions randomly sampled by their types. The questions are \colorbox{lightgray}{highlighted}.}
\label{tab:dataset_example}
\end{table*}

\section{Instruction Tuning}\label{appendix:instructiontuning}

Below is an instruction example used in the instruction tuning:

\begin{quote}
\texttt{"Below is an instruction that describes a task. Write a response that appropriately completes the request}\\
    \texttt{\textbf{Instruction}: Please answer the following questions concerning Azure Virtual Machines.} \\
    \texttt{\textbf{Input}: I have set Auto shut down time for my VM as 00:30 local time. I have updated the time on one day to 01:00 at 00:14. Though the modification took affect from next day. Is this expected?} \\
        \texttt{\textbf{Response}: Yes, this is the expected behavior. If you update the auto shutdown schedule for your VM within 30 minutes of the previously scheduled shutdown time, the new shutdown time takes effect the next day."} 
\end{quote}

\section{Training Setup and Parameters}\label{appendix:trainingparameter}

The source code, configurations, and data associated with our work can be accessed at: \url{https://aka.ms/Microsoft_QA}. Both the pre-training and instruction tuning are conducted in a V100 32GB node with 8 GPUs. The DeepSpeed\footnote{https://github.com/microsoft/DeepSpeed} framework was employed for the training process.

During the pre-training phase, we employed unsupervised learning with a next-word prediction training approach. We split the Azure documentation into 184,655 samples for pre-training. The pre-training parameters were set as follows: 8 training epochs, a maximum token length of 512, a batch size of 64, and a learning rate of $2e^{-5}$ with a cosine decaying scheduler. To ensure efficient processing, the Azure documentation was divided into separate samples, each with a maximum token length of 512 and no overlap. Image links and relative links to other Azure markdown files were removed, while clickable links were retained.

In the instruction-tuning phase, we selected QA pairs that had tags related to Azure, resulting in a dataset of 10,089 samples. To split the data into train and test sets, we computed the TF-IDF similarity between each pair of questions and excluded questions with high similarity from the test set. Consequently, the training set comprised 9,517 samples, while the test set contained 572 samples. We restrict the number of the test set considering the generation and evaluation cost with LLMs. The instruction tuning parameters were set as follows: 3 epochs, a maximum token length of 512, a batch size of 64, and a learning rate of $1e^{-5}$ with a cosine decaying scheduler. Note that we utilized a smaller number of epochs in the instruction-tuning process compared to pre-training to mitigate the risk of overfitting the training questions and answers.

\section{Baseline and Metric Prompts}\label{appendix:baselineprompt}
In this section, we list the prompts of baselines: LLM, LLM+BM25/DPR, and LLM+EXP from Table~\ref{tab:llmprompt} to Table~\ref{tab:llamagprompt}.

\begin{table}[!htb]
\small
\begin{tcolorbox}

[System]

As a helpful assistant, your task is to create responses to the user's questions. If you cannot be sure about the user's intention, please say, "Sorry, I do not understand your question"; If you cannot give a confident answer, please say, "Sorry, I cannot give a confident answer"\\

[User]

\textcolor{blue}{\textbraceleft question\textbraceright}

\end{tcolorbox}
\caption{The prompt of the raw \textit{LLM} method.}
\label{tab:llmprompt}
\end{table}

\begin{table}[!htb]
\small
\begin{tcolorbox}

[System]

As a helpful assistant, your task is to create responses to the user's questions. We have retrieved some chunks from the documents. These chunks are incomplete paragraphs and may not be relevant to the question. Please first determine whether these chunks are related to the user's question and disregard those you deem irrelevant. For the helpful chunks, integrate the useful content from these chunks into your answer without quoting them. If you cannot be sure about the user's intention, please say, "Sorry, I do not understand your question"; If you cannot give a confident answer, please say, "Sorry, I cannot give a confident answer". Below are the chunks:\\

<CHUNK>

\textcolor{blue}{\textbraceleft chunk 1\textbraceright}\\

<CHUNK>

\textcolor{blue}{\textbraceleft chunk 2\textbraceright}\\

<CHUNK>

\textcolor{blue}{\textbraceleft chunk 3\textbraceright}\\

[User]

\textcolor{blue}{\textbraceleft question\textbraceright}

\end{tcolorbox}
\caption{The prompt of the \textit{LLM+BM25/+DPR} method.}
\label{tab:retrievalprompt}
\end{table}

\begin{table}[!htb]
\small
\begin{tcolorbox}

[System]

As a helpful assistant, your task is to create responses to the user's questions. We have retrieved one response from another LLM. This answer may not be relevant to the question. If you think the LLM response is helpful, integrate the useful information into your answer without quoting them. Otherwise, you can ignore the LLM response. If you cannot be sure about the user's intention, please say, "Sorry, I do not understand your question"; If you cannot give a confident answer, please say, "Sorry, I cannot give a confident answer". Below are the LLM response:\\

<LLM RESPONSE>

\textcolor{blue}{\textbraceleft llama response\textbraceright}\\

[User]

\textcolor{blue}{\textbraceleft question\textbraceright}

\end{tcolorbox}
\caption{The prompt of the \textit{LLM+EXP} method.}
\label{tab:llamagprompt}
\end{table}

The prompt to extract keywords and spans in the KHR metric is shown in Table~\ref{tab:khr}. 

\begin{table}[!htb]
\small
\begin{tcolorbox}

[System]

 As a helpful assistant, your task is to extract the keywords or important spans from the provided text in <TEXT>. Focus on identifying significant words or phrases that are central to the topic or convey essential information. Take into account relevant context and consider both single words and multi-word expressions as potential keywords. Phrases follow the subject-verb or subject-verb-object pattern. The phrases should state if the verb is possible or not. Please provide a list of the extracted keywords or spans, separated by a comma. Below is the text: \\

[User]

<TEXT>: \textcolor{blue}{\textbraceleft grounded answer\textbraceright}\\

\end{tcolorbox}
\caption{The prompt to extract keywords and spans from the grounded answer in the KHR metric.}
\label{tab:khr}
\end{table}

\section{LLMs as Evaluators}\label{appendix:llmevaluator}

\subsection{Evaluator Prompt}\label{appendix:evaluatorprompt}
Table~\ref{tab:llm_evaluator} shows the prompt of scoring two responses. The LLM is tasked with comparing these responses to a grounded answer and providing evaluation explanations. Then LLM scores two responses ranging from 1 to 10.

\begin{table}[!htb]
\small
\begin{tcolorbox}

[System]

You are a helpful and precise assistant for checking the quality of the answer. We would like to invite you to provide feedback on the performance of two AI assistants in answering a user's question in \textless Question\textgreater, compared with the \textless Grounded Answer\textgreater written by humans. Please rate the helpfulness, relevance, accuracy, and level of detail of their responses. Each assistant receives an overall score on a scale of 1 to 10, where a higher score indicates better overall performance.\\

Please first provide a comprehensive explanation of your evaluation, avoiding any potential bias and ensuring that the order in which the responses were presented does not affect your judgment.\\

Then, output two lines indicating the scores for Assistant 1 and 2, respectively.\\

Output with the following format:\\
Evaluation evidence: \textless your evaluation explanation here\textgreater\\
Score of Assistant 1's response: \textless score\textgreater\\
Score of Assistant 2's response: \textless score\textgreater\\

[User]

\textless Question\textgreater: \textcolor{blue}{\textbraceleft question\textbraceright}\\
\textless Grounded Answer\textgreater: \textcolor{blue}{\textbraceleft grounded\_answer\textbraceright}\\
Assistant 1's Response: \textcolor{blue}{\textbraceleft response\_1\textbraceright}\\
Assistant 2's Response: \textcolor{blue}{\textbraceleft response\_2\textbraceright}

\end{tcolorbox}
\caption{The prompt of the LLM evaluator generates an evaluation explanation first and then gives scores on two response candidates.}
\label{tab:llm_evaluator}
\end{table}

\subsection{Evaluator Sensitivity}\label{appendix:evaluatorsensitity}
To evaluate the sensitivity of LLM evaluators to the positions of responses, we performed an experiment involving 200 randomly sampled response pairs from different methods. Each sample consisted of two responses from two different methods. We conducted two rounds of scoring by switching the positions of the responses. The responses were ranked based on their scores, and we assigned three tags: \textit{better}, \textit{tied}, and \textit{worse} to represent the ranking relationship. 
If the ranks are different in two rounds, we say there is a conflict. We observed scoring conflicts in the evaluations conducted by the LLM evaluator before and after switching the positions of the responses. In some cases, the LLM exhibited a preference for the response located in the first position, resulting in inconsistent rankings between the two rounds of scoring. 

\begin{figure}[!htb]
\centering
     \includegraphics[width=0.48\textwidth]{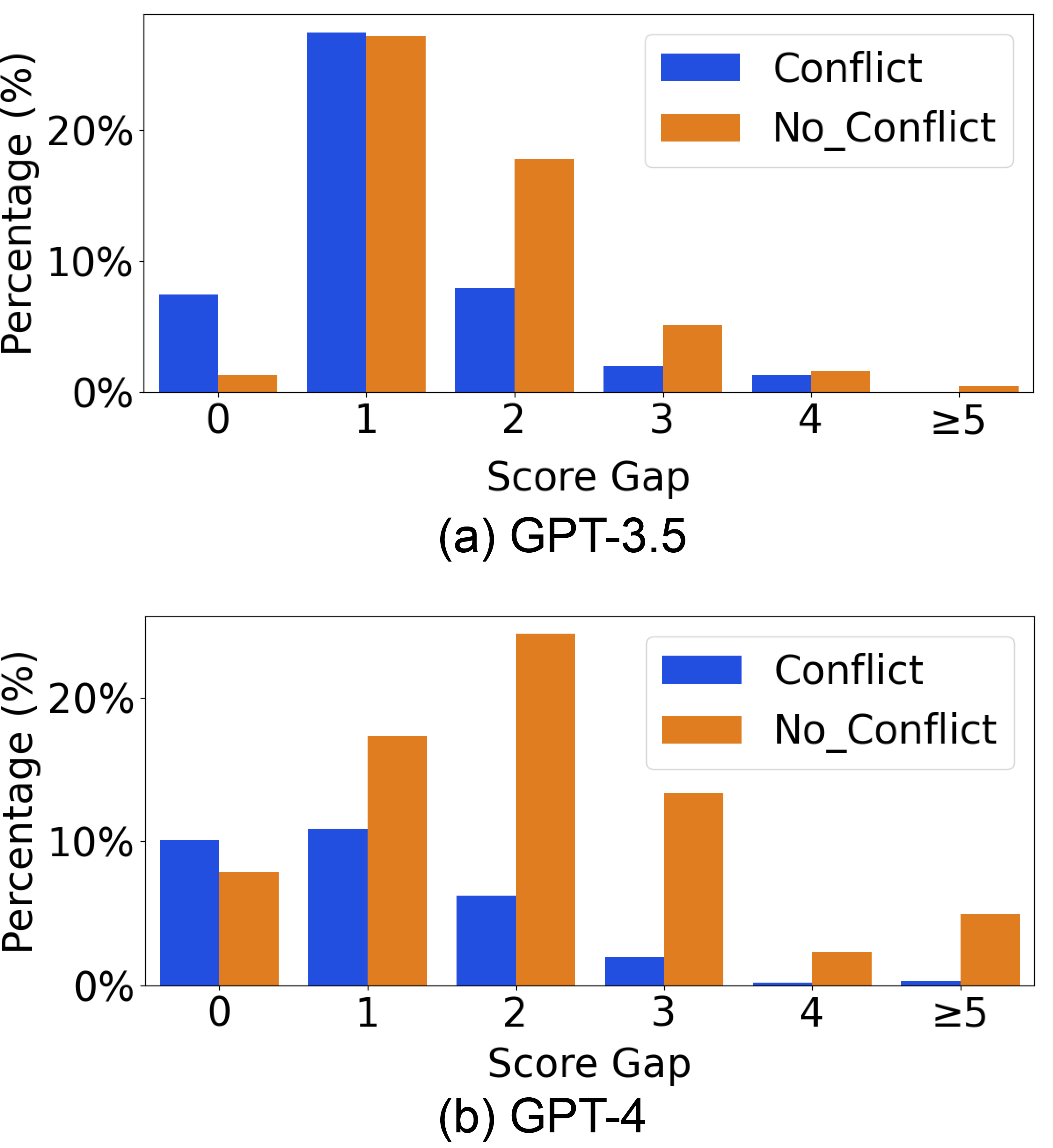}
     \caption{The statistics of score gaps. GPT-3.5 evaluator has 46.33\% conflict rate, and GPT-4 evaluator has 29.70\% conflict rate.}
     \label{fig:scoregap}
\end{figure}

\begin{table}[H]
\centering\resizebox{0.45\textwidth}{!}{
\begin{tabular}{l|cccccc}
\toprule
\multirow{2}{*}{\textbf{Evaluator}} & \multicolumn{6}{l}{\textbf{Conflict/Non-Conflict Ratio}}                    \\ \cline{2-7} 
                                    & \textbf{0} & \textbf{1} & \textbf{2} & \textbf{3} & \textbf{4} & \textbf{5} \\ \hline
\textbf{GPT-3.5}                    & 5.63       & 1.01       & 0.45       & 0.39       & 0.80       & 0          \\
\textbf{GPT-4}                      & 1.27       & 0.63       & 0.26       & 0.15       & 0.07       & 0.07       \\ \bottomrule
\end{tabular}}
\caption{The conflict versus non-conflict ratio of each score gap.}
\label{tab:conflictratio}
\end{table}

We introduce the concept of the score gap, which represents the absolute difference in scores between two responses within a single scoring round. Figure~\ref{fig:scoregap} shows the percentage of conflict and non-conflict cases when using GPT-3.5 and GPT-4 as evaluators, respectively. Note that each sample has two scoring rounds, and if a conflict arises between these two rounds, both rounds are labeled as \textit{conflict}. Notably, the GPT-4 evaluator exhibits a significantly lower conflict rate compared to GPT-3.5. Then, we select GPT-4 as our preferred evaluator. Furthermore, we observe that conflicts mostly occur within a score gap range of 0-2. On the other hand, we calculate the conflict/non-conflict ratio for each score gap value (see Table~\ref{tab:conflictratio}). When the score gap is 0 or 1, the ratio indicates a high probability of conflict. Based on these observations, we propose a rule where we trust the LLM evaluator only when the score gap exceeds 1. Otherwise, we conduct two scoring rounds by switching response positions and rank them based on the average score of two rounds. This approach mitigates scoring conflicts and ensures a reliable and efficient evaluation process, primarily relying on a single scoring round for most cases.

\section{Human Evaluation}\label{appendix:humanevaluation}

\subsection{Evaluation Setup and User Interface}\label{appendix:uiandsetup}

To ensure reliable evaluations, we randomly select a small subset of test samples consisting of 30 QA pairs. During the selection process, we exclude QA pairs that have grounded answers containing links or phrases such as \textit{"the answer is not supported in Microsoft Q\&A forum."} These types of grounded answers are not suitable for meaningful comparisons.

To conduct the evaluations, we engage crowdworkers who possess expertise in the cloud domain and are familiar with Microsoft products. We employ five such evaluators. Each sample receives five independent evaluations from these qualified evaluators. We consider an evaluation reliable when there is agreement among at least two out of the five evaluators. 

\begin{figure}[!htb]
\centering
     \includegraphics[width=0.48\textwidth]{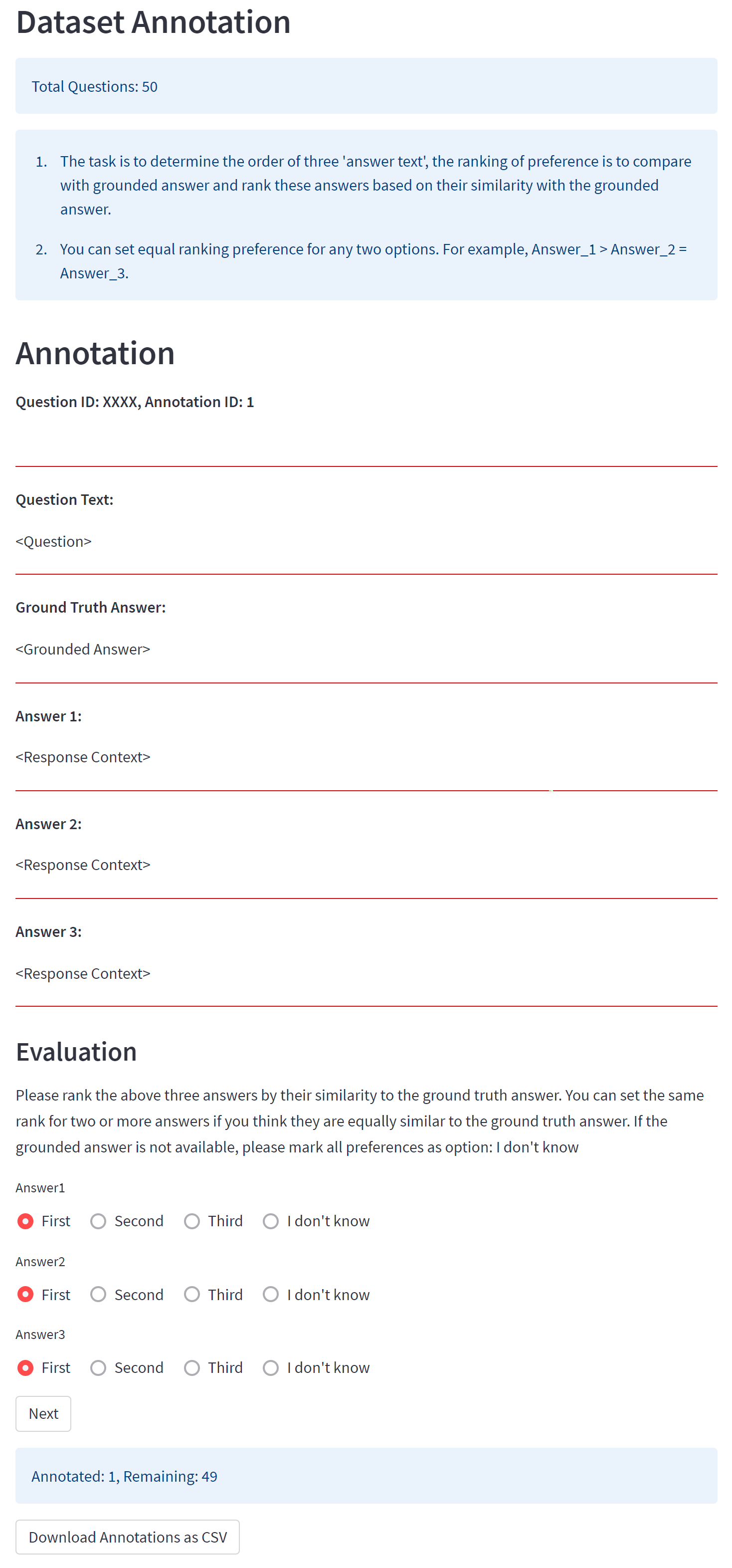}
     \caption{The user interface template of human evaluation. The details of QA and responses are not shown due to the space limit.}
     \label{fig:ui}
\end{figure}

Figure~\ref{fig:ui} shows the user interface (UI) of human evaluation in our study. The UI presents the components involved in evaluating a single sample. It begins with the display of a QA pair, followed by three responses generated by different methods. To minimize any potential bias, the positions of the responses are shuffled randomly for each sample evaluation. This ensures that the evaluator does not implicitly associate a particular response with a specific method. The evaluator's task is to rank the position of each response based on its similarity to the grounded answer. As the rank is assigned to each response individually, it is possible for two responses to receive the same rank. For example, both Response 1 and Response 2 can be assigned Rank 1 if they are equally similar to the grounded answer. This flexibility allows for a more nuanced evaluation and accommodates cases where multiple responses are equally relevant or accurate. The evaluators are also provided the \textit{"I don't know"} option if they do not have a confident evaluation of the sample. Before proceeding with the evaluation of the 30 test samples, each evaluator is given a separate test sample to familiarize themselves with the evaluation process. This preliminary test sample serves as a practice round, allowing the evaluators to become acquainted with the evaluation criteria and interface.

\subsection{Evaluation Agreements}\label{appendix:evaluationagreements}

\begin{figure}[!htb]
\centering
     \includegraphics[width=0.48\textwidth]{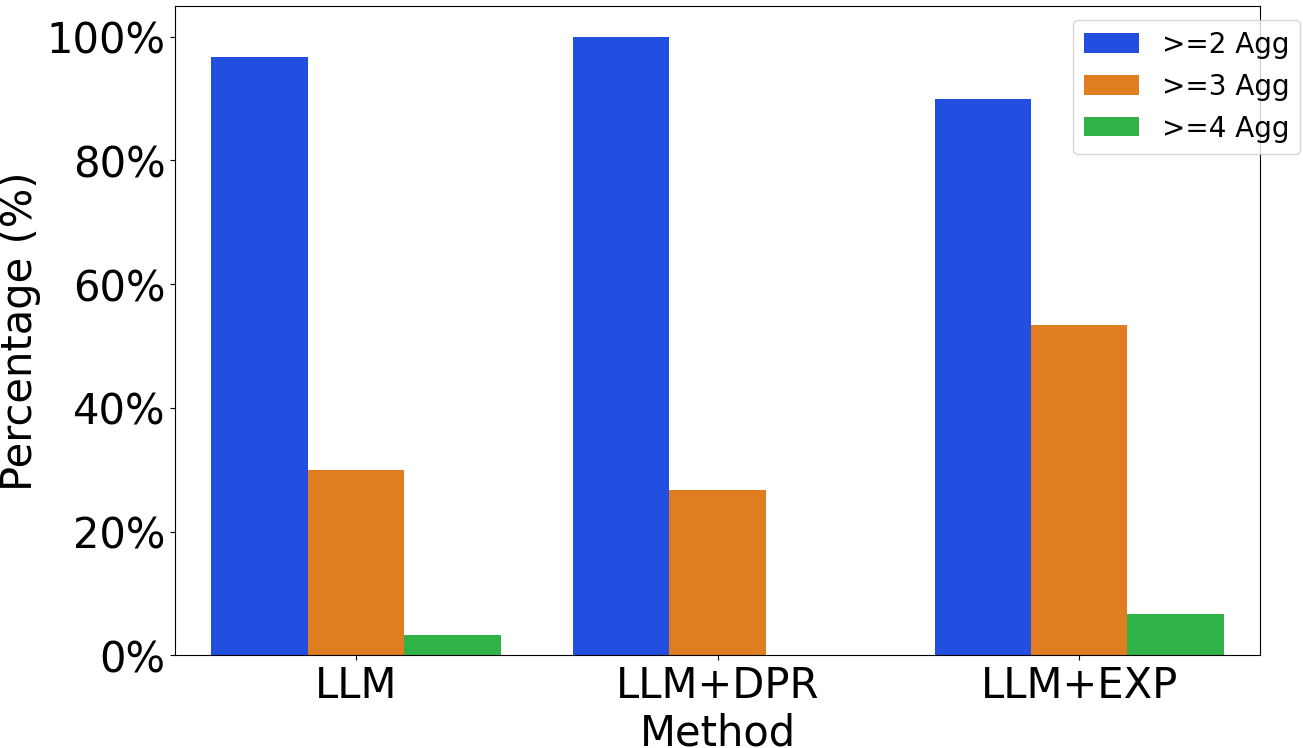}
     \caption{The statistics of agreements among human evaluators.}
     \label{fig:agreement}
\end{figure}

As shown in Figure~\ref{fig:agreement}, all evaluated methods consistently exhibit a nearly 100\% ratio of at-least-two-agreement. In particular, the LLM+EXP method stands out with a higher agreement compared to other approaches when considering agreement counts larger than 2. The results highlight the reliability of the human evaluation in achieving agreement across multiple annotations.

\section{Results of Lexical-overlap-based Metrics}\label{appendix:lexicalresults}

According to findings in \cite{krishna2021hurdles, ji2023survey}, as well as our own experimental observations, lexical-overlap-based metrics are inadequate for evaluating the quality of generated answers. This is evident from the results presented in Table~\ref{tab:lexicalmetric35} and Table~\ref{tab:lexicalmetric4}, where ROUGE scores demonstrate minimal variations across different methods. Although BLEU and METEOR indicate better performance for LLM+EXP, the differences are not significant. Additionally, the low values of BLEU and METEOR scores suggest that lexical-overlap-based metrics are not suited for comparing LLM-generated answers with human-written grounded answers.

\begin{table}[!htb]
\centering\resizebox{0.48\textwidth}{!}{
\begin{tabular}{l|cccc}
\toprule
\textbf{Metrics}   & \textbf{LLM} & \textbf{LLM+BM25} & \textbf{LLM+DPR} & \textbf{LLM+EXP} \\ \hline
\textbf{BLEU}                                               & 3.49   & 3.57          & 4.37        & \textbf{4.49}       \\
\textbf{ROUGE-1}                                            & 31.04  & \textbf{31.40}       & 31.31     & 30.49     \\
\textbf{ROUGE-2}                                            & 8.37  & 8.80         & \textbf{9.10}       & 8.63        \\
\textbf{ROUGE-L}                                            & 18.12 & \textbf{18.19}      & 18.02    & 17.92   \\
\textbf{METEOR}                                             & 17.77 & 18.07   & 20.50   & \textbf{20.67}    \\ \bottomrule
\end{tabular}
}
\caption{The results of lexical-overlap-based metrics over different methods with GPT-3.5 as the backbone LLM.}
\label{tab:lexicalmetric35}
\end{table}

\begin{table}[!htb]
\centering\resizebox{0.48\textwidth}{!}{
\begin{tabular}{l|cccc}
\toprule
\textbf{Metrics}   & \textbf{LLM} & \textbf{LLM+BM25} & \textbf{LLM+DPR} & \textbf{LLM+EXP} \\ \hline
\textbf{BLEU}                                               &  3.78    & 3.74         &  4.64        & \textbf{5.55}        \\
\textbf{ROUGE-1}                                            & 30.62  & \textbf{32.62}       &  31.39      & 31.62     \\
\textbf{ROUGE-2}                                            & 8.87    & \textbf{9.92}         &  9.67        & 9.29        \\
\textbf{ROUGE-L}                                            & 17.37  & \textbf{18.76}       & 18.12      & 18.34      \\
\textbf{METEOR}                                             &  23.03  & 22.02  & 22.83   & \textbf{23.63}    \\ \bottomrule
\end{tabular}
}
\caption{The results of lexical-overlap-based metrics over different methods with GPT-4 as the backbone LLM.}
\label{tab:lexicalmetric4}
\end{table}

\section{Case Study}\label{appendix:casestudy}

We present two case studies to offer a detailed comparison of different methods. 

Table~\ref{tab:frontcasedetail} presents a question inquiring about whether cloud-only users can log on to a hybrid joined computer. The grounded answer is negative, as only users with both on-prem and cloud presence can log on to hybrid AAD. We compare the results obtained from retrieved chunks and our domain knowledge generated from the domain-specific LM. The data-retrieved chunks contain scattered information related to keywords in the question, such as AAD joined devices and configurations of hybrid AAD joined devices. However, they do not directly address the question of whether cloud-only users can log on to hybrid AAD joined devices. On the other hand, our domain knowledge provides a direct answer, stating that users require an on-prem account to log on to hybrid joined devices. Incorporating this extra information results in significantly different responses, with data-retrieval+LLM suggesting it is a configuration issue.

Table~\ref{tab:case1} presents another question asking about obtaining metrics for specific blob containers' transactions and the affected files. The grounded answer recommends using Log Analytics to aggregate statistics from logs, a method not available in the Azure portal. The retrieved chunks gathered information on Azure Storage and Azure Monitor Metrics but did not offer a clear approach for obtaining specific metrics for blob containers. Consequently, data-retrieval+LLM responses suggested that the metrics could be obtained through the Azure portal and erroneously mentioned Azure Storage Analytics as a possible solution. On the other hand, our response from the domain-specific LM explicitly suggests utilizing Azure Log Analytics and provides a step-by-step guideline, aligning with the grounded answer.

\onecolumn
\begin{longtable}{p{53pt}| p{360pt}}
\hline
\multirow{3}{*}{Question} & Cloud-only account on hybrid joined device?\\
&Hello everyone,  \\
&we are planning to use hybrid joined devices (AD \& AAD) in future as one step (of many) to a cloud-only approach. \\
&What we are currently wondering (because it's not working in the lab environment):  \\
&Can a cloud-only user logon to a hybrid joined computer? Currently in our test environment it's not working. The company want's to have the On-Premise AD User Accounts removed in near future and use cloud-only accounts.  \\
&Thanks in advance \\ \hline
\multirow{3}{*}{\begin{tabular}[c]{@{}l@{}}Grounded \\ Answer\end{tabular}} 
& If the machine is **Hybrid AAD joined,** a cloud only user wont be able to join, as while you login to a Hybrid AAD join machine, usually the user is present in both the on-prem and the cloud and only that user can login, because while logging into a Domain joined machine, the user first has to get authenticated by the on-prem **Domain Controller** before AAD checks the credentials. Second option is to go ahead with **Azure AD Joined** machine. \\\hline
\multirow{3}{*}{\begin{tabular}[c]{@{}l@{}}BM25 \\ Chunks\end{tabular}}   & Deploy Azure AD joined VMs in Azure Virtual Desktop-Azure. User accounts can be cloud-only or synced users from the same Azure AD tenant. \#\# Known limitations The following known limitations may affect access to your on-premises or Active Directory domain $\cdots$\\ \cline{2-2}
& Archive for What's new in Azure Active Directory? Users can now use FIDO2 security keys to sign in to their Hybrid Azure AD joined Windows 10 devices and get seamless sign-on to their on-premises and cloud resources. $\cdots$\\\cline{2-2}
 & What is an Azure AD joined device? | Azure AD Join | Description | | - - - | - - - | | * * Definition * * | Joined only to Azure AD requiring organizational account to sign in to the device $\cdots$ \\ \hline
\multirow{3}{*}{\begin{tabular}[c]{@{}l@{}}DPR \\ Chunks\end{tabular}} 
& Some hybrid solutions might require access to both on-premises and cloud resources. An example of a use case would be an Identity Governance solution that uses a service account on premises for access to AD DS and requires access to Azure AD. $\cdots$\\  \cline{2-2}
&- **Hybrid Azure AD joined device** - Users might be able to use the device to sign-in to their on-premises domain. However, they can't access Azure AD resources such as Microsoft 365. $\cdots$\\  \cline{2-2}
&without a hybrid deployment that includes an on-premises AD DS environment. The following table outlines common device ownership models and how they would typically be joined to a domain: $\cdots$\\ \hline
\multirow{2}{*}{\begin{tabular}[c]{@{}l@{}}Our Domain \\ Knowledge\end{tabular}}
& A cloud-only account can't logon to a hybrid joined computer. You need to have an on-premises AD account to logon to a hybrid joined computer.
\\ \hline
\multirow{1}{*}{\begin{tabular}[c]{@{}l@{}}LLM\end{tabular}}
&Sorry, I cannot give a confident answer." \\\hline
\multirow{3}{*}{\begin{tabular}[c]{@{}l@{}}LLM+BM25\end{tabular}}
&Yes, a cloud-only user can log on to a hybrid joined computer. However, since you mentioned that it's not working in your lab environment, there might be some configuration issues or limitations that need to be addressed.\\
&Ensure that your hybrid environment is properly set up, and the devices are correctly joined to both on-premises Active Directory and Azure Active Directory. Additionally, make sure that the user accounts are properly synced between the on-premises AD and Azure AD.\\
&If you still face issues, it's recommended to review your hybrid setup and troubleshoot any potential configuration problems.\\ \hline
\multirow{3}{*}{\begin{tabular}[c]{@{}l@{}}LLM+DPR\end{tabular}}
&Yes, a cloud-only user can log on to a hybrid joined computer. However, if it's not working in your test environment, there might be some configuration issues or requirements that need to be addressed.\\
&To enable cloud-only users to log on to hybrid joined devices, you need to ensure the following:\\
&1. The devices must be running Windows 10 or later.\\
&2. The devices must be joined to both on-premises Active Directory Domain Services (AD DS) and Azure Active Directory (Azure AD).\\
&3. Azure AD Connect must be configured and running to synchronize your on-premises AD DS with Azure AD.\\
&4. Seamless Single Sign-On (SSO) should be enabled to allow users to access Azure AD resources without needing to re-enter their credentials.\\
&If you have met these requirements and are still experiencing issues, you may need to troubleshoot your configuration or check for any errors in the synchronization process.\\ \hline
\multirow{3}{*}{\begin{tabular}[c]{@{}l@{}}LLM+EXP\end{tabular}}
&No, a cloud-only user cannot logon to a hybrid joined computer. To logon to a hybrid joined computer, you need to have an on-premises AD account. If your company plans to remove on-premises AD user accounts in the near future, you may need to consider other options such as using Azure AD Domain Services or implementing a password hash synchronization solution. These options will allow you to continue using cloud-only accounts while still being able to logon to hybrid joined devices.\\ \hline
\caption{
Case details on the front page. GPT-4 is employed as the backbone LLM generator. The question and grounded answer are presented, followed by summarized BM25 and DPR chunks, along with domain knowledge derived from our domain-specific LM. Due to space limitations, only the top chunks of BM25 and DPR are displayed. Then we present responses from four baselines: LLM, LLM+BM25, LLM+DPR, LLM+EXP. The question and answer is found in: \url{https://learn.microsoft.com/en-us/answers/questions/2096/}
}\label{tab:frontcasedetail}\\
\end{longtable}

\begin{longtable}{p{53pt}| p{360pt}}
\hline
\multirow{3}{*}{Question} 
& Get metrics on Azure Storage Account Blob being used, not the entire data store.\\
&Is it possible via the Azure Portal to get metrics on specific BLOB containers' transactions and the files they affected? I can only get metrics for the entire storage account and multiple apps/services use the same account, I need to find out which container is being used so I can narrow down which app consumes the most resources on the storage account. \\ \hline
\multirow{3}{*}{\begin{tabular}[c]{@{}l@{}}Grounded \\ Answer\end{tabular}} 
&With Log Analytics, the measurement can be achieved by aggregating storage logs on container/folder. You can refer to the below for detailed guidelines. \\
&You need to calculate based on request or response sizes. Here's an example for Ingress/Egress:\\
&\begin{lstlisting}
StorageBlobLogs     
| extend Ingress = iff(isnull(RequestHeaderSize),0,RequestHeaderSize) + iff(isnull(RequestBodySize),0,RequestBodySize)    
| extend Egress = iff(isnull(ResponseHeaderSize),0,ResponseHeaderSize) + iff(isnull(ResponseBodySize),0,ResponseBodySize)     
| project OperationName, Ingress, Egress, RequestHeaderSize, RequestBodySize, ResponseHeaderSize, ResponseBodySize
\end{lstlisting} \\
& Azure Monitor   Metrics: \url{https://learn.microsoft.com/en-us/azure/storage/common/storage-metrics-in-azure-monitor}    \\
&If the above suggestion doesn't help you: The workaround is to aggregate the statistics from analytics logs to capture changes plus create a capacity baseline by enumerating objects in each container. Will you please elaborate on what usage of containers you need? What's the model of container layout versus the final users? How many containers and blobs in each container can be projected? Kindly let us know if the above helps or you need further assistance on this issue.\\ \hline
\multirow{3}{*}{\begin{tabular}[c]{@{}l@{}}BM25 \\ Chunks\end{tabular}}   
& To understand how disallowing public access may affect client applications, we recommend that you enable logging and metrics for that account and analyze patterns of anonymous requests over an interval of time. Use metrics $\cdots$\\ \cline{2-2}
& List the account-level metric definition You can list the metric definition of your storage account or the Blob storage service. Use the [ az monitor metrics list - definitions ] command.  $\cdots$\\\cline{2-2}
 & - [ Can I enable persistence on a previously created cache? ] - [ Can I enable AOF and RDB persistence at the same time? ] - [ How does persistence work with geo-replication? ] - [ Which persistence model should I choose? ] - $\cdots$ \\ \hline
\multirow{3}{*}{\begin{tabular}[c]{@{}l@{}}DPR \\ Chunks\end{tabular}} 
& Azure Storage is a core Azure service, and forms an important part of the majority of solutions that customers deploy to the Azure infrastructure. Azure Storage includes capabilities $ \cdots$\\  \cline{2-2}
&To log Azure Storage data with Azure Monitor and analyze it with Azure Log Analytics, you must first create a diagnostic setting that indicates what types of requests and for which storage services you want to log data. To create $\cdots$\\  \cline{2-2}
&Enable blob inventory reports for your storage account. See [Enable Azure Storage blob inventory reports](blob-inventory-how-to.md). Use the following  $\cdots$\\ \hline
\multirow{2}{*}{\begin{tabular}[c]{@{}l@{}}Our Domain \\ Knowledge\end{tabular}}
&There is no way to get container level metrics from the portal or the metrics service. You would need to write some code to do this. If you are using Azure Monitor(\url{https://learn.microsoft.com/en-us/azure/azure-monitor/overview?WT.mc_id=AZ-MVP-5000120}) you can create a diagnostic setting to send the logs and metrics for Blob storage to an Azure Log Analytics workspace. From there you can query the logs to get the information you need.
\\ \hline
\multirow{1}{*}{\begin{tabular}[c]{@{}l@{}}LLM\end{tabular}}
& Yes, it is possible to get metrics on specific Blob containers' transactions and the files they affected in Azure Portal. You can use the Azure Storage Analytics feature to collect metrics and logs for your Blob containers. \\
&To enable Storage Analytics, you need to go to your Storage account in the Azure Portal, select "Storage Analytics" from the left-hand menu, and then configure the settings for metrics and logs. You can choose to enable metrics for specific Blob containers and set the retention period for the data. Once you have enabled Storage Analytics, you can view the metrics and logs in the Azure Portal by going to your Storage account, selecting "Metrics" from the left-hand menu, and then selecting the Blob container and time range you want to view. This will give you detailed information on the transactions and files affected by that container.\\\hline
\multirow{3}{*}{\begin{tabular}[c]{@{}l@{}}LLM+BM25\end{tabular}}
&Yes, it is possible to get metrics on specific Blob containers' transactions and the files they affected via the Azure Portal. You can use Azure Metrics Explorer to track anonymous requests to a storage account and determine which containers are being accessed anonymously. To do this, you can enable logging and metrics for that account and analyze patterns of anonymous requests over an interval of time. Use metrics to determine the number of anonymous requests to the storage account, and use logs to determine which containers are being accessed anonymously. Additionally, you can list the metric definition of your storage account or the Blob storage service using the \textit{az monitor metrics list-definitions} command. This will allow you to see the metrics for specific Blob containers. I hope this helps!\\ \hline
\multirow{3}{*}{\begin{tabular}[c]{@{}l@{}}LLM+DPR\end{tabular}}
&Yes, it is possible to get metrics on specific Blob containers' transactions and the files they affected. You can use Azure Storage Analytics to monitor and analyze the usage of your Blob containers. \\
&To get metrics on a specific Blob container, you can follow these steps:\\
&1. Enable Storage Analytics for your Blob container. You can do this by navigating to your Blob container in the Azure portal, selecting "Monitoring" from the left-hand menu, and then selecting "Storage Analytics" and enabling it. \\
&2. Once Storage Analytics is enabled, you can view metrics for your Blob container by selecting "Metrics" from the left-hand menu. From there, you can select the Blob container you want to view metrics for and choose the specific metrics you want to see, such as "Transactions" or "Egress".\\ \hline
\multirow{3}{*}{\begin{tabular}[c]{@{}l@{}}LLM+EXP\end{tabular}}
&Unfortunately, you cannot directly get container-level metrics from the Azure Portal or the metrics service. However, you can achieve this by using Azure Monitor and Azure Log Analytics workspace.\\
&To get metrics on specific Blob containers' transactions and the files they affected, follow these steps:\\
&1. Set up Azure Monitor \url{https://docs.microsoft.com/en-us/azure/azure-monitor/overview} for your storage account.\\
&2. Create a diagnostic setting to send the logs and metrics for Blob storage to an Azure Log Analytics workspace.\\
&3. Query the logs in the Log Analytics workspace to get the information you need.\\
&By querying the logs in the Log Analytics workspace, you can filter the data based on container names and analyze the transactions and files affected. This will help you identify which container is being used the most and narrow down the app consuming the most resources on the storage account.\\ \hline
\caption{
Case study on Azure Storage metrics. GPT-4 is employed as the backbone LLM generator. The question and grounded answer are presented, followed by summarized BM25 and DPR chunks, along with domain knowledge derived from our domain-specific LM. Due to space limitations, only the top chunks of BM25 and DPR are displayed. Then we present responses from four baselines: LLM, LLM+BM25, LLM+DPR, LLM+EXP. The question and answer is found in: \url{https://learn.microsoft.com/en-us/answers/questions/172078/}
}\label{tab:case1}\\
\end{longtable}

\end{document}